\documentclass[10pt,twocolumn,letterpaper]{article}

\usepackage{cvpr}
\usepackage{times}
\usepackage{epsfig}
\usepackage{graphicx}
\usepackage{amsmath}
\usepackage{amssymb}

\usepackage[normalem]{ulem}

\usepackage{newtxtext}

\usepackage{booktabs}
\usepackage{multirow, makecell}

\usepackage[pagebackref=true,breaklinks=true,letterpaper=true,colorlinks,bookmarks=false]{hyperref}

 \cvprfinalcopy 

\def\cvprPaperID{5611} 

\ifcvprfinal\fi
\begin{document}

\title{AdaBits: Neural Network Quantization with Adaptive Bit-Widths}

\author{Qing Jin\thanks{Equal Contribution. Work is done when Qing Jin is an intern at ByteDance.}\qquad\qquad\qquad\qquad\quad Linjie Yang\footnotemark[1]\qquad\qquad\qquad\qquad\quad Zhenyu Liao\\
{\tt\small jinqingking@gmail.com}\qquad\qquad {\tt\small linjie.yang@bytedance.com}\qquad\qquad {\tt\small liaozhenyu2004@gmail.com}\\
ByteDance Inc.}

\maketitle

\begin{abstract}
Deep neural networks with adaptive configurations have gained increasing attention due to the instant and flexible deployment of these models on platforms with different resource budgets. In this paper, we investigate a novel option to achieve this goal by enabling adaptive bit-widths of weights and activations in the model. We first examine the benefits and challenges of training quantized model with adaptive bit-widths, and then experiment with several approaches including direct adaptation, progressive training and joint training. We discover that joint training is able to produce comparable performance on the adaptive model as individual models. We also propose a new technique named Switchable Clipping Level (S-CL) to further improve quantized models at the lowest bit-width. With our proposed techniques applied on a bunch of models including MobileNet V1/V2 and ResNet50, we demonstrate that bit-width of weights and activations is a new option for adaptively executable deep neural networks, offering a distinct opportunity for improved accuracy-efficiency trade-off as well as instant adaptation according to the platform constraints in real-world applications.
\end{abstract}

\section{Introduction}
Recent development of deep learning enables application of deep neural networks across a wide range of platforms that present different resource constraints. For example, popular mobile apps such as TikTok and Snapchat on portable devices pose stringent requirements on response latency and energy consumption, while visual recognition system embedded in a self-driving vehicle~\cite{grigorescu2019survey,teichman2011practical} is more demanding on fast and accurate prediction. Besides, for medical application~\cite{Li2019, zhou2019hyper} applied with portable testing systems, implementing efficient model will accelerate the diagnosing process to save time for doctor and patient. The problem is more serious if other factors are taken into account, such as aging of hardware, battery conditions, as well as different versions of software systems. To serve applications under all these scenarios with drastically different requirements, different models tailored for different resource budgets can be devised either manually~\cite{he2016deep,he2016identity,howard2017mobilenets,sandler2018mobilenetv2} or automatically through neural architecture search~\cite{tan2019mnasnet,zoph2016neural,zoph2018learning}. This strategy is beneficial for optimal trade-offs with a fixed combination of constraints, but is not economical, because it requires time-consuming training and benchmarking for each of these models, which prohibits instant adaptation to favor different scenarios. To tackle this problem, recent work focuses on training a single model that is flexible and scalable. For example, \cite{yu2018slimmable} proposes a method where the number of channels can be adjusted through changing the width-multiplier in each layer. Inspired by this work, \cite{cai2019once} integrates adaptation of depth, width and kernel size altogether, and achieves better trade-offs between performance and efficiency through progressive training.~\cite{scalingup} adopts the same strategy with scaling up factors, but uses simultaneous training algorithm to achieve improved predictive accuracy.

\begin{figure*}[h!]
\begin{center}
   \includegraphics[width=0.8\linewidth]{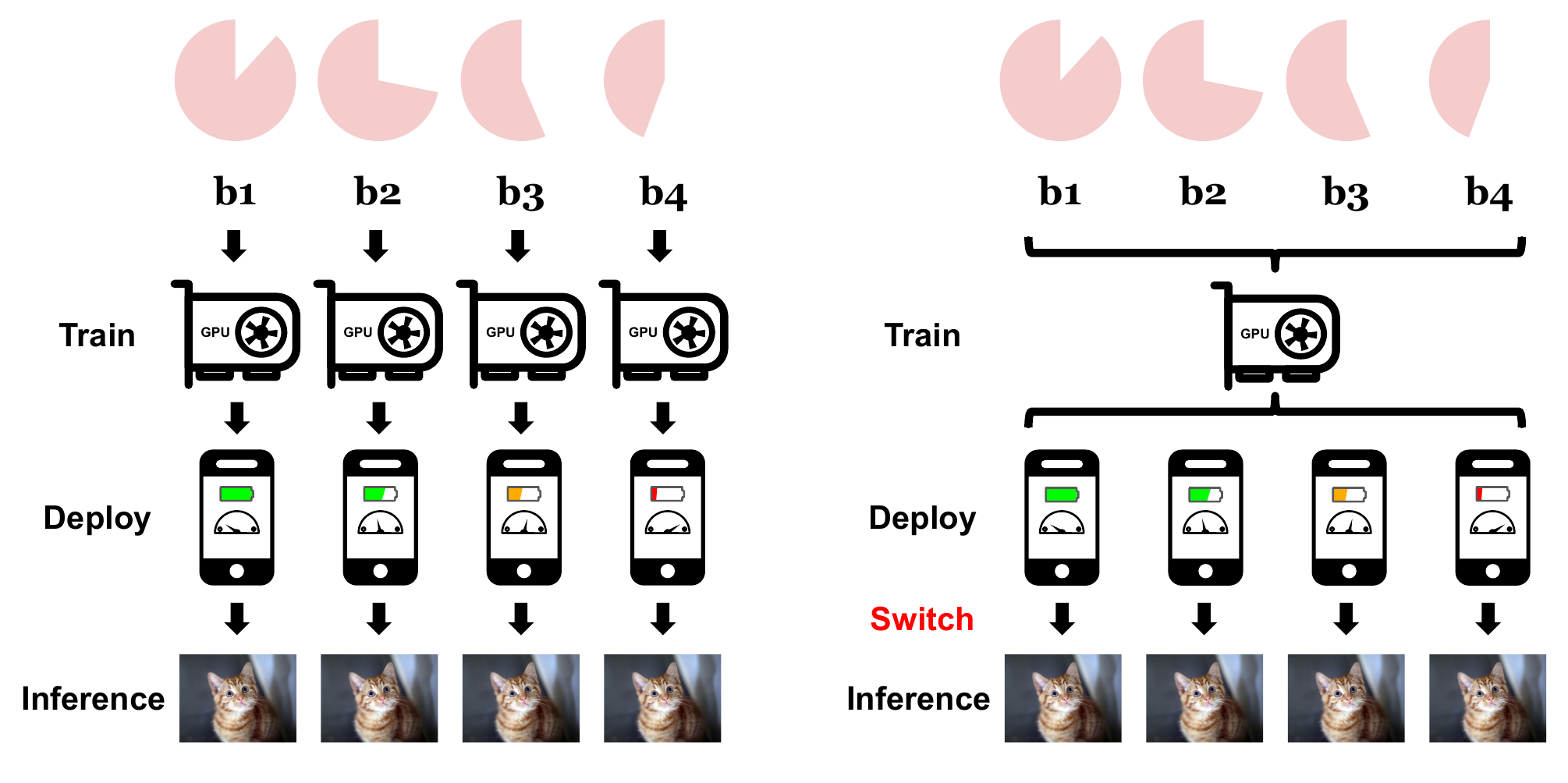}
\end{center}
\vspace{-15pt}
   \caption{Deployment of neural networks with different bit-widths according to the computational budget. Left: Individually train several quantized models with different bit-widths for each scenario. Right: Train a single model quantized with adpative bit-widths and switch to the proper bit-width in real application based on the device condition.}
\label{fig:adabits}
\end{figure*}

Surprisingly, albeit the above-mentioned methods achieve the desired flexibility of adaptive deployment, bit-width of weights and intermediate activations, as another degree of freedom, is almost overlooked in previous work. Suppose we can adaptively choose bit-width for a neural network during inference without further training, it will provide an distinct opportunity for more powerful model compression and acceleration. As an example, compared with model with full-precision, quantizing MobileNet V2 to 6-bit compresses the model size by roughly $4.74\times$ and reduces the BitOPs by $14.25\times$\footnote{According to the IEEE Standard 754, floating-point number is represented with 23-bits mantissa, and here we simplify the analysis by approximating the effective BitOPs of floating-point multiplication with 23-bit fixed-point multiplication.}, while scaling the model's channel numbers by a width-multiplier of $0.35\times$ only shrinks the model size by $2.06\times$ and cuts down the FLOPs by $5.10\times$. Moreover, as presented in~\cite{sat}, 6-bit MobileNet V2 demonstrates improved predictive capability than the full-precision counterpart, while reducing channel numbers to $0.35\times$ will significantly impair its performance~\cite{sandler2018mobilenetv2}. It becomes more contrastive if other constraints are taken into account, such as memory cost, latency and energy consumption. Additionally, adaptive bit-widths is generally applicable to most key building blocks of deep neural networks, including time-consuming convolutional and fully-connected layers. Meanwhile, adaptive deployment will also introduce negligible computation, as discussed in~\cite{yu2018slimmable}. Figure~\ref{fig:adabits} illustrates the basic concept of quantization with adaptive bit-widths.

At the first glance, adaptive bit-widths might be trivial and handy, as weights and activations with different precisions may not differ from each other very much. If so, model trained under some specific precision will be able to directly provide good performance under other bit-widths. However, as we will see in the following, such na\"ive method is not applicable, because important information will be lost during shrinkage or enlargement of bit-widths in the neural network. Even more deliberate method of progressively training quantized models with different bit-width(s) fails to achieve the optimal performance, as the finetuning process sabotages important property of the model, thus significantly diminishes the validation accuracy of the model when quantized back to the original bit-width.

All the above evidence indicates that quantization with adaptive bit-widths does not come as free lunch as it might appear, but is more subtle and involves new mechanisms that require meticulously designed techniques. In this work, we try to investigate this topic, and study specific methods to train quantized neural networks adaptive to different requirements. We utilize the state-of-the-art learning-based quantization method of Scale-Adjusted Training~\cite{sat} as a baseline scheme for individual-precision quantization. We find that an adaptive model produced by a joint quantization approach with a key treatment to the clipping level parameters~\cite{choi2018pact} is able to achieve comparable performance with individual-precision models on several bit-widths. The treatment to the clipping levels is named \emph{Switchable Clipping Level (S-CL)}. S-CL accommodates large activation values for high-precision quantization, and prevents undesired increasing of clipping levels for low-precision cases. Through some empirical analysis, we find that unnecessarily large clipping levels might cause large quantization error, and impact the performance of quantized model, especially on the lowest precision. To our best knowledge, this work is the first to tackle this problem of producing quantized models with adaptive bit-widths.

This paper is organized as following. After summarizing some related works in Section~\ref{sec:related_work}, we first revisit the recent work of scale-adjusted training (SAT)~\cite{sat}, which is adopted in our whole study. In Section~\ref{sec:main}, we first illustrate potential benefits and challenges of quantization with adaptive bit-widths. Then we propose a joint training approach with a new technique named switchable clipping level based on the analysis of some baseline results. In Section~\ref{sec:experiments}, we show that with the proposed techniques, the adaptive models could achieve comparable accuracies as individual ones on different bit-widths and for a wide range of models including MobileNet V1/V2 and ResNet50.


\section{Related Work}
\label{sec:related_work}

{\bf Neural Network Quantization} Neural network quantization has long been studied since the very beginning of the recent blooming era of deep learning, including binarization~\cite{bai2018proxquant,courbariaux2015binaryconnect,courbariaux2016binarized,rastegari2016xnor}, quantization~\cite{li2016ternary,zhou2016dorefa,zhu2016trained} and ensemble method~\cite{zhu2019binary}. Initially, uniform precision quantization is adopted inside the whole network, where all layers share the same bit-width~\cite{jacob2018quantization,leng2018extremely,mellempudi2017ternary,mishra2017apprentice,mishra2017wrpn,park2017weighted,xu2018alternating,zhou2017balanced}. Recent work employs neural architecture search methods for model quantization, which implements mixed-precision strategy where different bit-widths are assigned to different layers or even channels~\cite{elthakeb2018releq, lou2019autoqb,uhlich2019differentiable,wang2019haq,wu2018mixed}.~\cite{sat} analyzes the problem of efficient training for neural network quantization, and proposes a scale-adjusted training (SAT) technique, achieving state-of-the-art performance. However, the possibility of developing a single model applicable at different bit-widths is still not well-examined, and it remains unclear how to achieve this purpose.

{\bf Neural Architecture Search} Neural architecture search (NAS) gains increasing popularity in recent study~\cite{proxylessnas,li2020autonl,progressivenas,mei2020atomnas,enas,fbnet,snas,zoph2016neural}. Specifically, the searching strategy is adopted in other aspects of optimizing neural networks, such as automatic tuning of various training hyper-parameters including activation function~\cite{ramachandran2018searching} and data augmentation~\cite{cubuk2018autoaugment}. The NAS algorithms also benefit other tasks, such as generative adversarial networks~\cite{gong2019autogan}, object detection~\cite{chen2019detnas} and segmentation~\cite{liu2019auto}. As mentioned above, neural architecture search method for quantization is also actively studied in recent literature. However, NAS is computationally expensive, and usually requires time-consuming re-training or finetuning. Recent work has reduced the searching time by a large extent through one-shot architecture search~\cite{bender2018understanding,stamoulis2019single}. However, the resulting models are still inflexible, prohibiting their application in adaptive scenarios. Generally, conventional NAS methods are more suitable for optimizing a single model under specific resource constraints.

{\bf Adaptive neural networks} Different from but related to NAS,~\cite{yu2018slimmable} proposes to simultaneously train a single model with different width multipliers, to achieve instant adaptation for different application requirements. Following this line,~\cite{cai2019once} explores adjustment of width, depth and kernel sizes simultaneously, achieving better predictive accuracy under the same computational constraints through progressive training.~\cite{scalingup} extends similar strategy to large-size models, and further employs a NAS algorithm to discover better models. However, these methods neglect the option of quantization with different bit-widths in their strategies, leaving quantization with adaptive bit-widths an open problem.

\section{Revisiting Scale-Adjusted Training (SAT)}
Quantization usually comes with performance degeneration, as the model capacity is significantly reduced in comparison with the full-precision counterpart. However, a recent study~\cite{sat} demonstrates that a large portion of accuracy degradation is caused by inefficient training where learning-based quantization, potentially acting as a regularization, actually provides an opportunity to improve generalization capability. The key idea is that quantized models usually enforce large variance in their weights, which brings about over-fitting issue during training. Based on this finding,~\cite{sat} proposes a simple yet effective method, called scale-adjusted training (SAT), which scales the weights down to a healthy level for network optimization. Specifically, constant scaling is applied to the quantized weights of linear layers without BN by
\begin{equation}
\label{eq:quant_constant_rescale}
    Q^*_{ij}=\frac{1}{\sqrt{n_{\mathrm{out}}\mathbb{VAR}[Q_{rs}]}}Q_{ij}
\end{equation}
where $Q_{ij}$ is a quantized weight and $n_{\mathrm{out}}$ is the number of output neurons in this layer. By combining with a quantization approach named parameterized clipping activation (PACT)~\cite{choi2018pact}, SAT facilitates more efficient training, enabling quantized models to perform consistently and significantly better than conventional quantization techniques, sometimes even surpassing their full-precision counterparts. 
Due to the numerous algorithms for neural network quantization, it is difficult, if not impossible, to experiment with different quantization algorithms for the adaptive bit-widths problem. To this end, we adopt the PACT algorithm with the SAT technique, which gives the state-of-the-art performance for neural network quantization, throughout all of our experiments. For brevity, we refer to this approach as SAT.

\section{Quantization with Adaptive Bit-widths}
\label{sec:main}
In this section, we first examine the benefits and challenges of quantization with adaptive bit-widths. We explore direct adaptation and progressive quantization as two straightforward methods towards this goal with unsatisfying results. We then propose a novel joint quantization approach to deal with the challenge and achieve the same level performance with the adaptive models compared to the individual models.

\subsection{Benefit and Challenges}

\begin{figure*}[h!]
\begin{center}
   \includegraphics[width=0.35\linewidth]{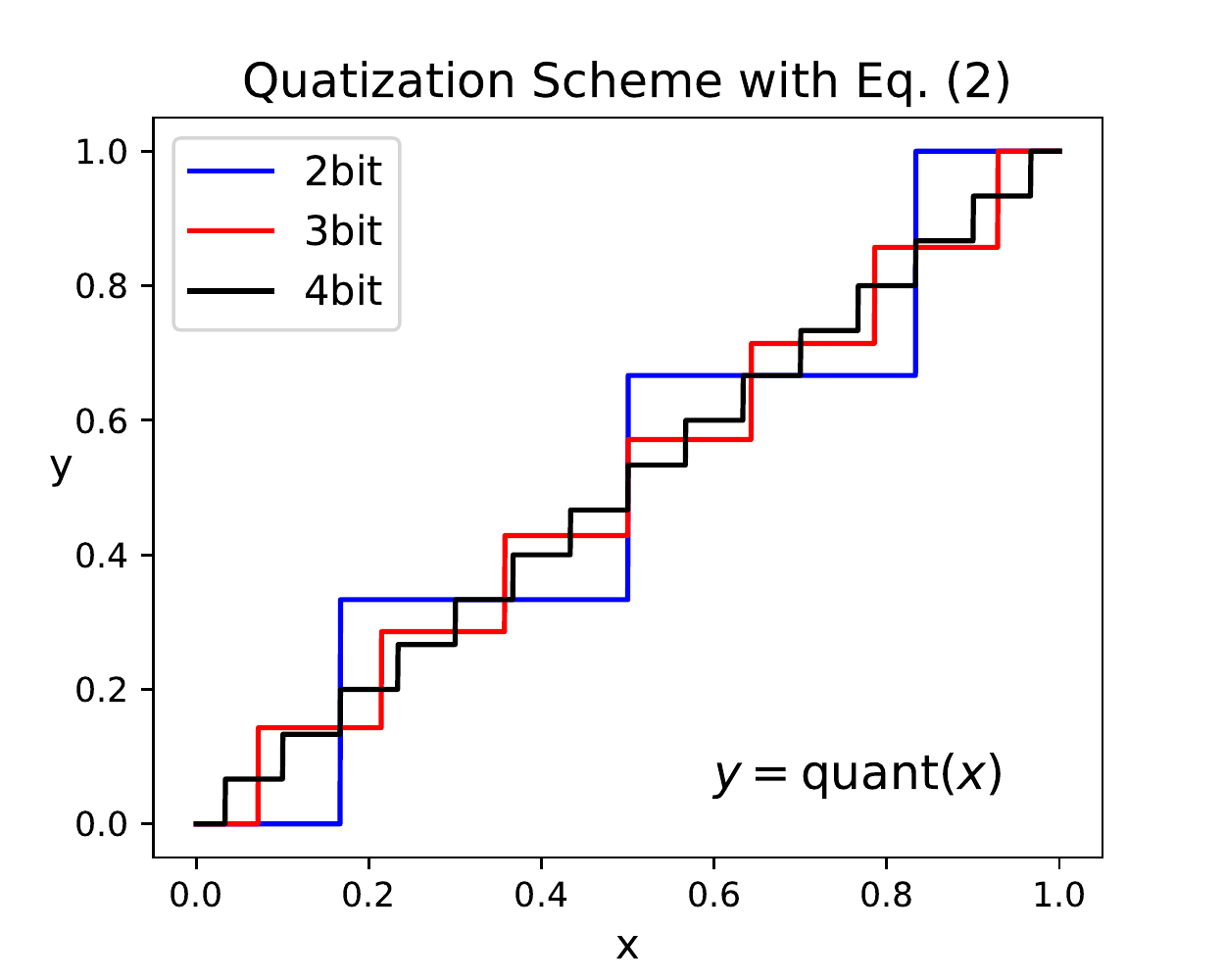}
   \quad
   \includegraphics[width=0.35\linewidth]{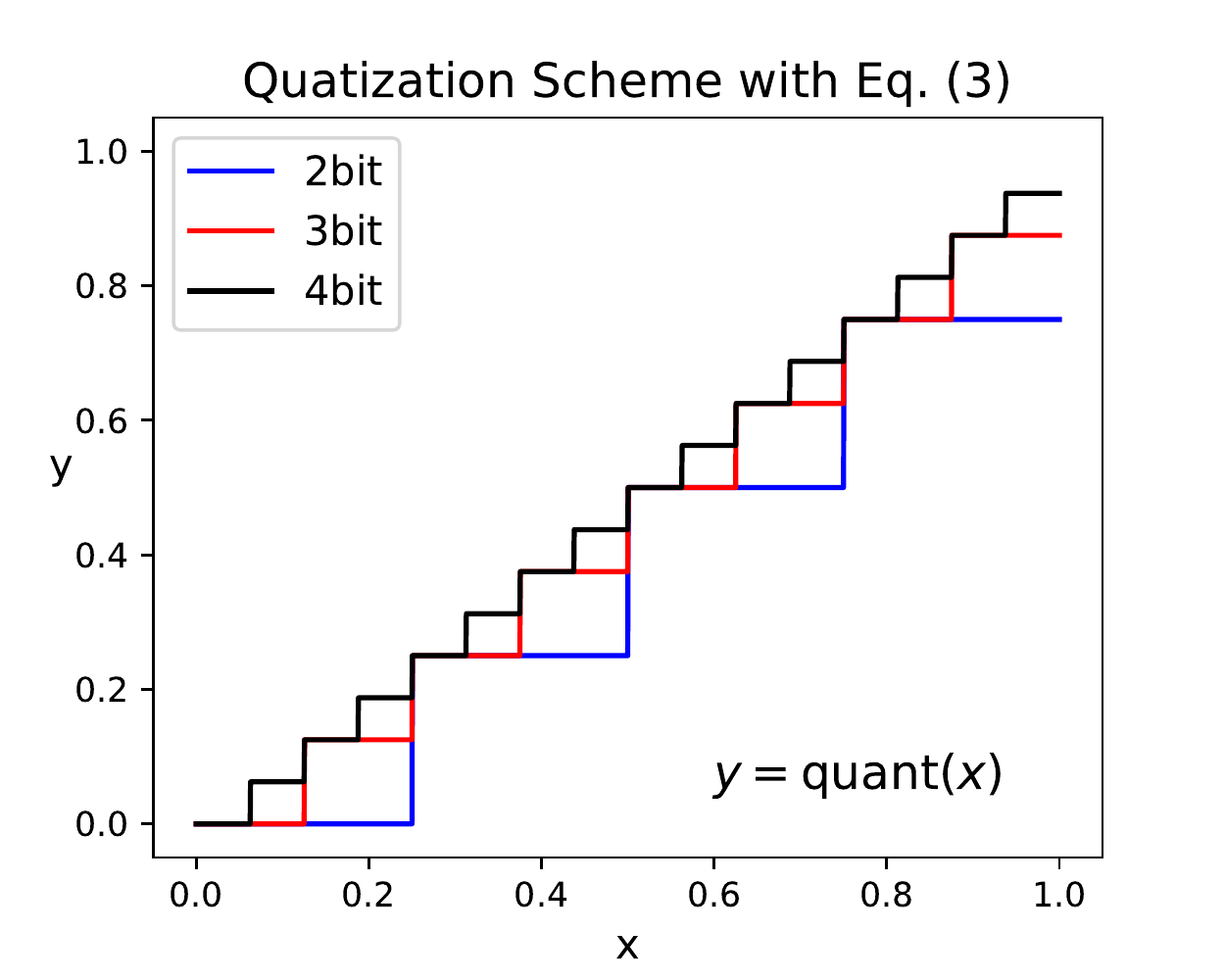}
\end{center}
\vspace{-10pt}
   \caption{Comparison of two quantization schemes: original scheme (Eq.~\eqref{eq:quant}) and modified scheme (Eq.~\eqref{eq:quant_modified}).}
\label{fig:scheme}
\vspace{-10pt}
\end{figure*}

Neural network quantization provides significant reduction in model size, latency, and energy consumption. Training a single quantized models executable at different bit-widths poses a great opportunity for flexible and adaptive deployment since models with larger bit-width are still consistently better than those with smaller bit-width. Actually, for MobileNet V1/V2, changing the bit-width from 4bit to 8bit can enlarge the model size by $1.7\times$ and the BitOPs by $3.2\times$, while the predictive accuracy can change by $1.5\%$ on the ImageNet dataset with SAT~\cite{sat}. From this we can see that there is a noticeable trade-off between accuracy and efficiency on quantized models. In the following, we will first investigate two straightforward methods for adaptive bit-widths, which will reveal some key challenges of this problem.

\subsubsection{Modified DoReFa Scheme} \label{sec:modified}
Before more detailed analysis, we would like to emphasize a distinct difficulty encountered in quantized models with adaptive bit-widths. The DoReFa scheme~\cite{zhou2016dorefa} is adopted in the original SAT method for weight quantization, where weights are quantized with
\begin{equation}
\label{eq:quant}
    q_k(x)=\frac{1}{a}\Big\lfloor ax\Big\rceil
\end{equation}
Here, $\lfloor\cdot\rceil$ indicates rounding to the nearest integer, and $a$ equals $2^k-1$ where $k$ is the number of quantization bits. However, as illustrated in Figure~\ref{fig:scheme}, such a scheme is not practical for quantization with adaptive bit-width, as there is no direct mapping between weights quantized to different bit-widths, disabling direct conversion of quantized models from a bit-width to lower bit-widths. It necessitates storage of the full-precision weights, and the quantization procedure needs to be repeated for different bit-widths during model deployment. This significantly increases the size of the stored model, and greatly limits the applications of the model. To accommodate simple conversion of quantized models, we modify the scheme to use a quantization function given by
\begin{equation}
\label{eq:quant_modified}
    q_k(x)=\frac{1}{\widehat{a}}\mathrm{min}\Big(\Big\lfloor \widehat{a}x\Big\rfloor, \widehat{a}-1\Big)
\end{equation}
Here, $\lfloor\cdot\rfloor$ indicates the floor rounding function, and $\widehat{a}$ equals $2^k$ where $k$ is the number of quantization bits. This quantizing function does not differ quite much from that for the original DoReFa scheme, and should give similar performance for quantized model. Moreover, as shown in Figure~\ref{fig:scheme}, it enables direct adaptation from higher bit-width to lower bit-width through discarding lower bits in the weights directly. We formulate this capacity with the following theorem which can be easily proved.

{\noindent\bf Theorem 1} For any $x$ in $[0, 1]$ and any two positive integers $a > b$,
\begin{equation}
\label{eq:quant_modified_proof}
    \lfloor2^ax\rfloor>>(a-b)=\lfloor2^bx\rfloor
\end{equation}

 In the following, we first utilize the original DoReFa scheme to explore quantization with adaptive bit-widths, and compare with the SAT method~\cite{sat}. More experiment results will be provided using both the original scheme and the modified scheme in Section~\ref{sec:experiments}.

\subsubsection{Direct Adaptation}

\begin{figure*}[h!]
\begin{center}
   \includegraphics[width=0.35\linewidth]{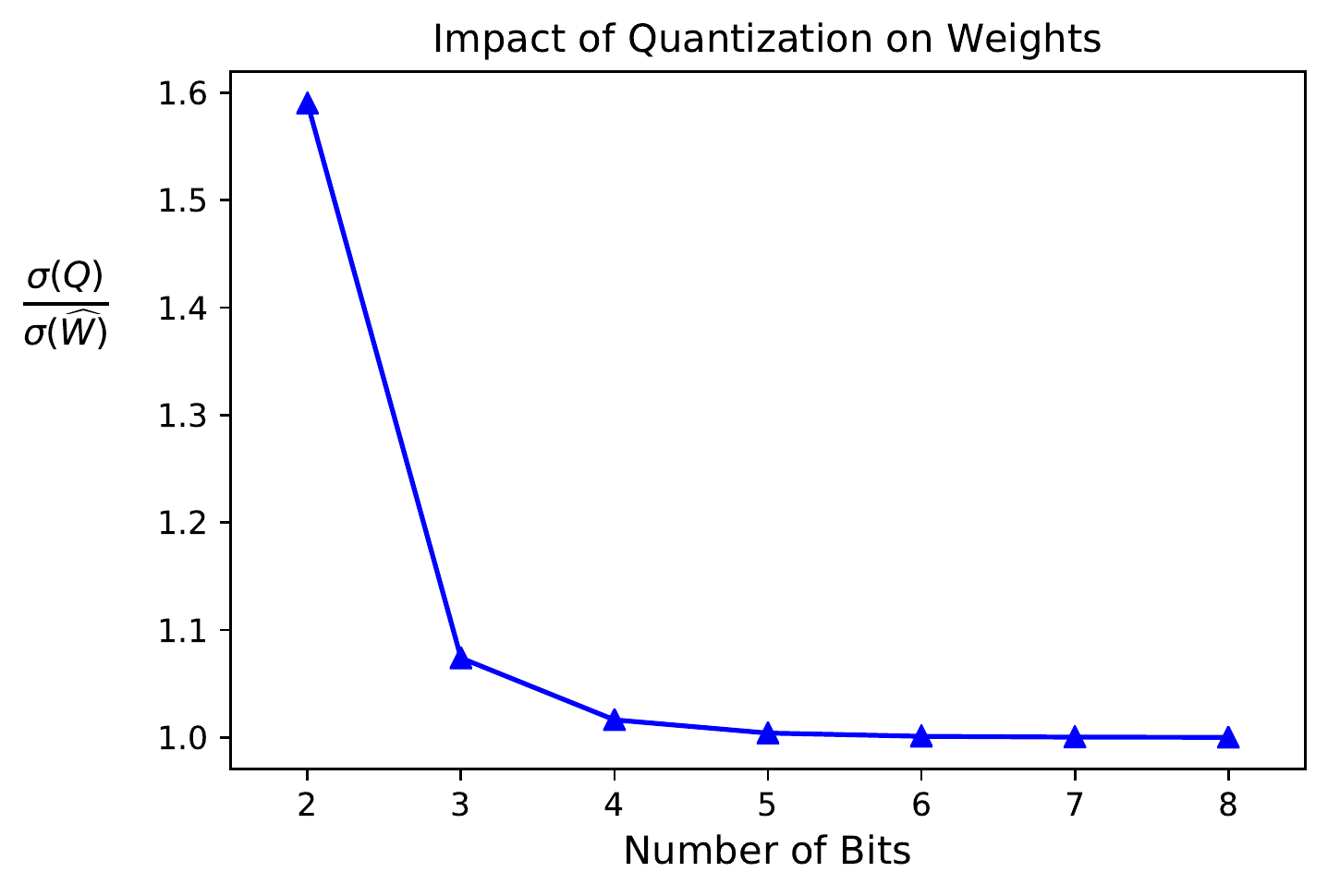}\quad
   \includegraphics[width=0.35\linewidth]{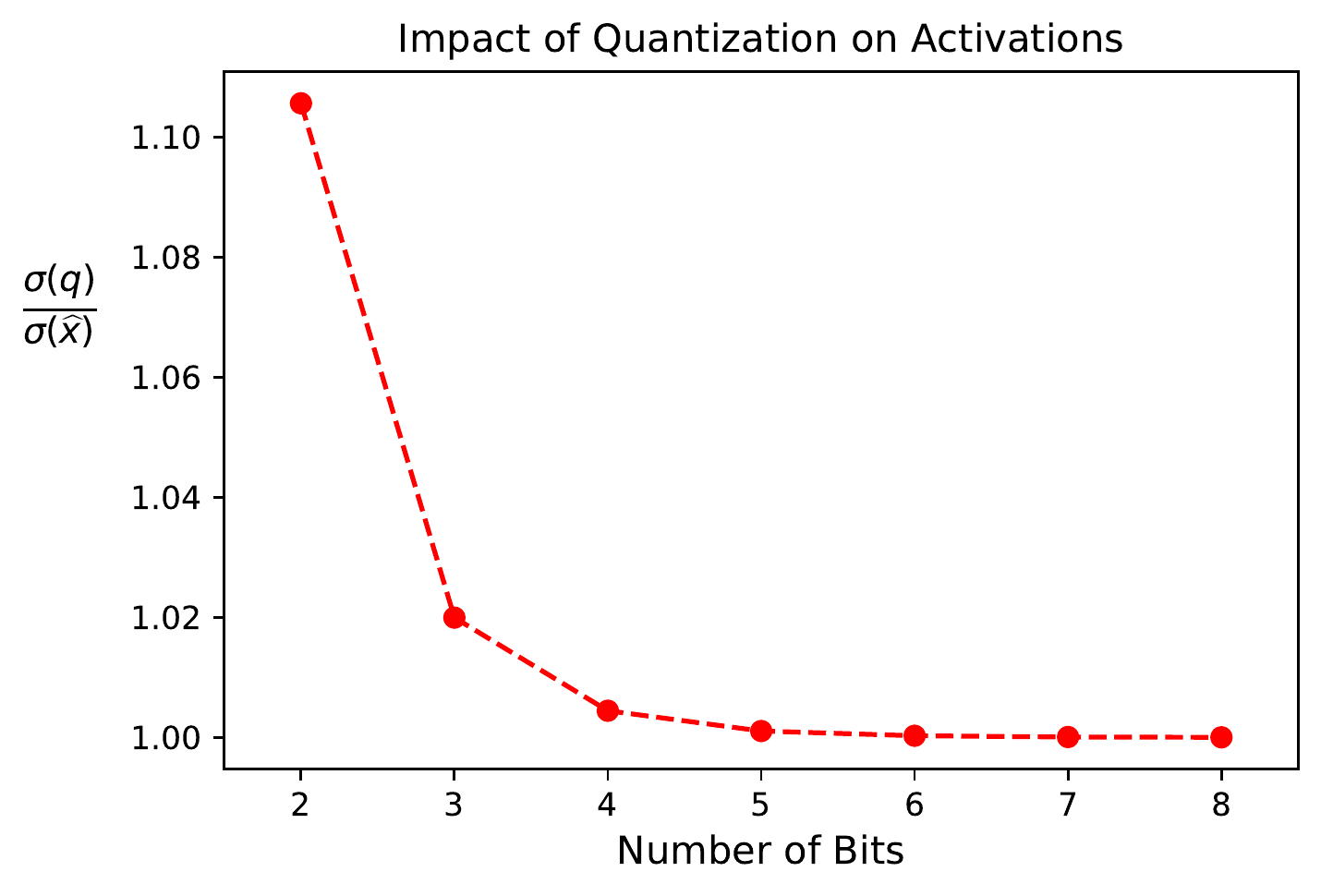}
\end{center}
   \caption{Impact of quantization on variances of weights and activation under different bit-widths. The variance of both quantized weights and activations gets larger when the bit-wdith gets smaller.}
\label{fig:stddev}
\end{figure*}

We first investigate whether quantized models trained on one bit-width can be directly used on other bit-widths. This cheap approach could be viable since weights with different bit-widths might be close to each other in value. To check if this method is practical, we evaluate the validation accuracy of ResNet50 on ImageNet by adjusting the bit-width to several different settings, where the original weights are trained under either the lowest or the highest bit-width (2 bit and 4 bit in this case, respectively). As indicated in previous research~\cite{sat}, quantization with different bit-widths entails difference in variances of weights and activations, as illustrated in Figure~\ref{fig:stddev}. Thus the networks trained on one bit-width suffer from the mismatch of the layer statistics when evaluated on another bit-width. To alleviate this problem, we apply batch norm (BN) calibration introduced in~\cite{autoslim} to calibrate the statistics in batch normalization layers for reasonable comparison.

The results with and without BN calibration are listed in Table~\ref{tab:naive_method}, together with performance of models trained and evaluated under the same bit-width using SAT. It is shown that without BN calibration, models trained on one bit degenerate significantly on another bit. With BN calibration, model trained on 2 bit successfully preserves the performance on larger bits, but is still inferior to results achieved by directly training on the large bits; also, model trained on 4 bit degenerates severely on smaller bits.
In summary, models trained and evaluated in different bit-widths are not suitable for adaptive deployment of quantization models due to the difference in training and evaluation settings. Particularly, models trained with larger bit-width suffer from more serious performance degeneration when quantized to lower precision, while training with smaller bit-width limits the potential of models deployed on higher precision. 

\begin{table}[h!]
\begin{center}
\scalebox{.9}{
\begin{tabular}{l|cccc}
\hline
Model & 4 bit & 3 bit & 2 bit \\
\hline\hline
4 bit Trained (wo/ BN calib) & 76.3 & 67.1 & 0.3 \\
2 bit Trained (wo/ BN calib) & 41.1 & 48.7 & 73.3 \\
\hline
4 bit Trained (w/ BN calib) & 76.3 & 73.2 & 20.3 \\
2 bit Trained (w/ BN calib) & 73.4 & 73.2 & 73.3 \\
\hline
SAT~\cite{sat} & 76.3 & 75.9 & 73.3 \\
\hline
\end{tabular}
}
\end{center}
\caption{Direct adaptation of models trained on 2 and 4 bits on different bit-widths, with and without batch norm calibration. Results are top-1 validation accuracy (\%) of ResNet50 on ImageNet.}
\label{tab:naive_method}
\end{table}

\subsubsection{Progressive Quantization}
The above analysis demonstrates that quantization with adaptive bit-widths is not directly available from models trained with individual bit-width. In this section, we investigate the possibility of progressive training, where a quantized model is trained on multiple bit-widths sequentially. It can be conducted in two ways, where the bit-width for training can be gradually increased or decreased. We experiment with both methods for ResNet50 on ImageNet. For either case, we use the model trained individually with the highest (lowest) bit-width as the initial point, which is finetuned with the second highest (lowest) bit-width, and further finetuned with the next bit-width. We continue this finetuning process until all bit-widths under consideration are processed. For each phase of finetuning, the same hyper-parameters are adopted as those for training individual quantization. Also, BN calibration is applied to the final model on different bit-widths for reasonable comparison. The results are shown in Table~\ref{tab:progressive}.

In Table~\ref{tab:progressive}, the model first trained with 2 bit and fine-tuned with ascending bit-width achieves good result at the final 4 bit, but is corrupted on lower bits. The model first trained with 4 bit and fine-tuned with descending bit-width only achieve slightly better performance than directly applying 2 bit model on multiple bits in Table~\ref{tab:naive_method}, which does not preserve performance of the higher 3 and 4 bits. The above results indicate that progressive training might have introduced undesired perturbation to the model trained previously, which impairs its original performance. This shows the progressive training method is still not suitable for models with adaptive bit-widths.

\begin{table}
\begin{center}
\scalebox{.9}{
\begin{tabular}{l|ccc}
\hline
Model & 4 bit & 3 bit & 2 bit \\
\hline\hline
Ascending Bit-width & 76.3 & 73.4 & 29.5 \\
Descending Bit-width & 73.9 & 73.6 & 73.5 \\
SAT~\cite{sat} & 76.3 & 75.9 & 73.3 \\
\hline
\end{tabular}
}
\end{center}
\caption{Results of progressive quantization with ascending/descending bit-widths of ResNet50 on ImageNet. Results are top-1 validation accuracy (\%).}
\label{tab:progressive}
\end{table}

\subsection{Joint Quantization}

The above results show that sequential training does not preserve the model characteristics in previously trained bit-widths, which indicates that the model weights for different bit-widths should be jointly optimized. Specifically, we adopt a joint training approach similar to slimmable neural networks~\cite{yu2018slimmable}. Instead of training models with different channel numbers, we simultaneously train models under different bit-widths with shared weights. Also, as mentioned above, quantization with different bit-widths leads to different variances of quantized weights and activations. Based on this, we adopt the switchable batch normalization technique introduced in~\cite{yu2018slimmable}. We call this method \emph{Vanilla AdaBits}, and the performance is listed in Table~\ref{tab:adabits}. It can be seen that models trained with all of the bit-widths achieve comparable performance as those trained individually, which validates the effectiveness of this approach. However, there is still a performance gap for the lowest bit-width at 4 bit, which is $0.5\%$ lower than the individually trained model. This is undesired and further improvement needs to be made.

\begin{table}
\begin{center}
\scalebox{.9}{
\begin{tabular}{l|ccccc}
\hline
Model & 8 bit & 6 bit & 5 bit & 4 bit\\
\hline\hline
Vanilla AdaBits & 72.4 & 72.5 & 72.1 & 70.8 \\
SAT~\cite{sat} & 72.6 & 72.3 & 71.9 & 71.3 \\
\hline
\end{tabular}
}
\end{center}
\caption{ Results of Vanilla AdaBits with MobileNet V1 on ImageNet with four bit-widths. Results are top-1 validation accuracy (\%).}
\label{tab:adabits}
\end{table}

\begin{figure}[h!]
\begin{center}
   \includegraphics[width=0.7\linewidth]{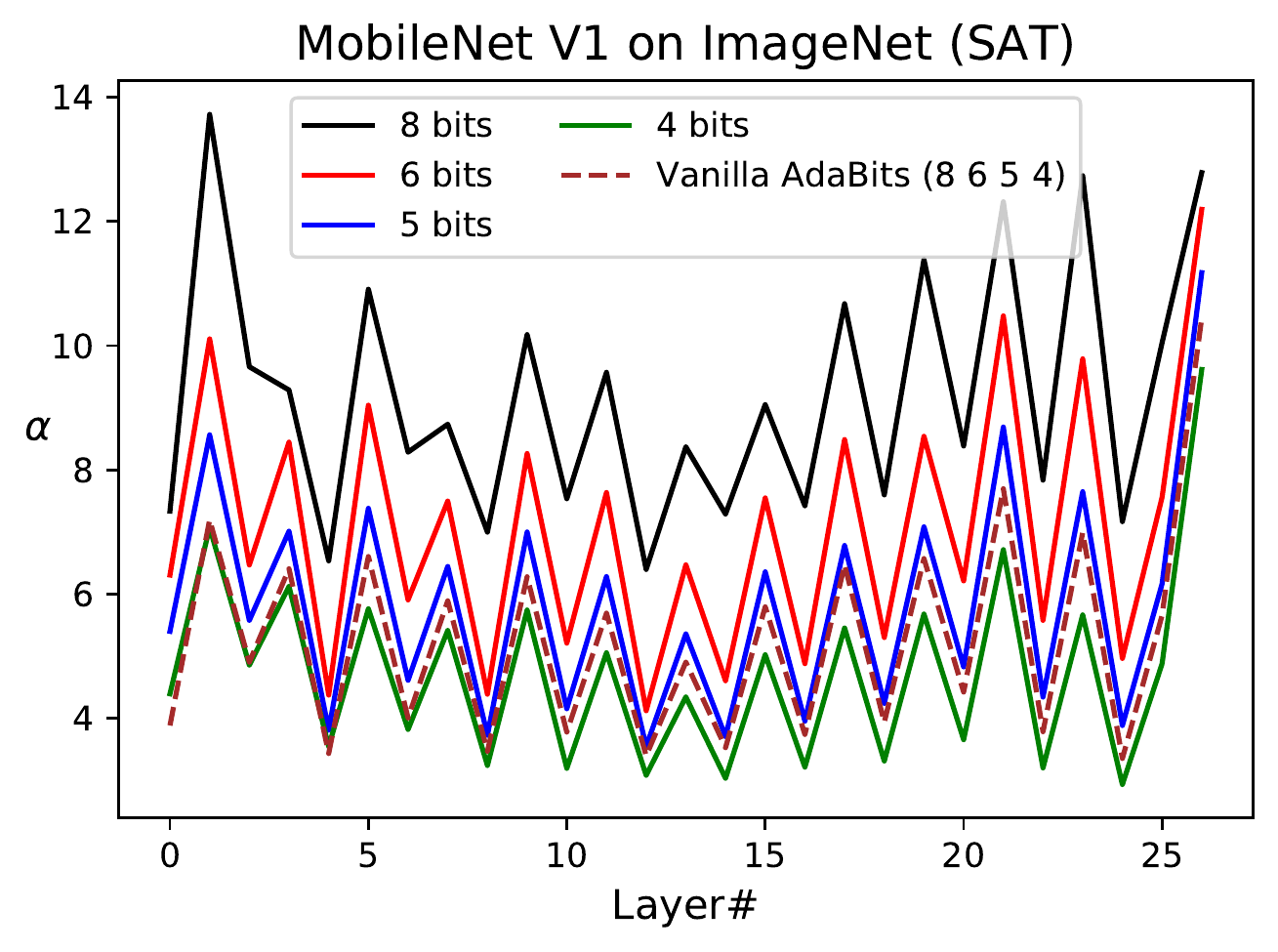}
\end{center}
   \caption{Clipping levels in different layers for models trained individually with different bit-widths (solid lines) or trained with Vanilla AdaBits (dashed line). Note that the clipping level of a layer refers to the clipping level for the output of this layer. The outputs of the last layer are not clipped.}
\label{fig:clipping_levels_sat}
\end{figure}

In the PACT algorithm adopted by SAT, the activations of each layer will be first clipped by a learned parameter $\alpha$ named \emph{clipping level}, and then quantized to discrete numbers. Specifically, an activation value $x$ is first clipped to the interval $[0,\alpha]$, then scaled, quantized and rescaled to produce the quantized value $q$ as
\begin{subequations}
\begin{align}
\label{eq:clip}
    \widetilde{x}&=\frac{1}{2}\Big[|x|-|x-\alpha|+\alpha\Big]\\
    q&=\alpha q_k\Big(\frac{\widetilde{x}}{\alpha}\Big)
\end{align}
\end{subequations}
Note that in the original paper of PACT~\cite{choi2018pact}, the authors found that different bit-widths result in different clipping levels. In the Vanilla AdaBits, the clipping levels of different bit-widths are shared, which may potentially disturb the optimization process of the network.

\begin{figure}[h!]
\begin{center}
   \includegraphics[width=0.7\linewidth]{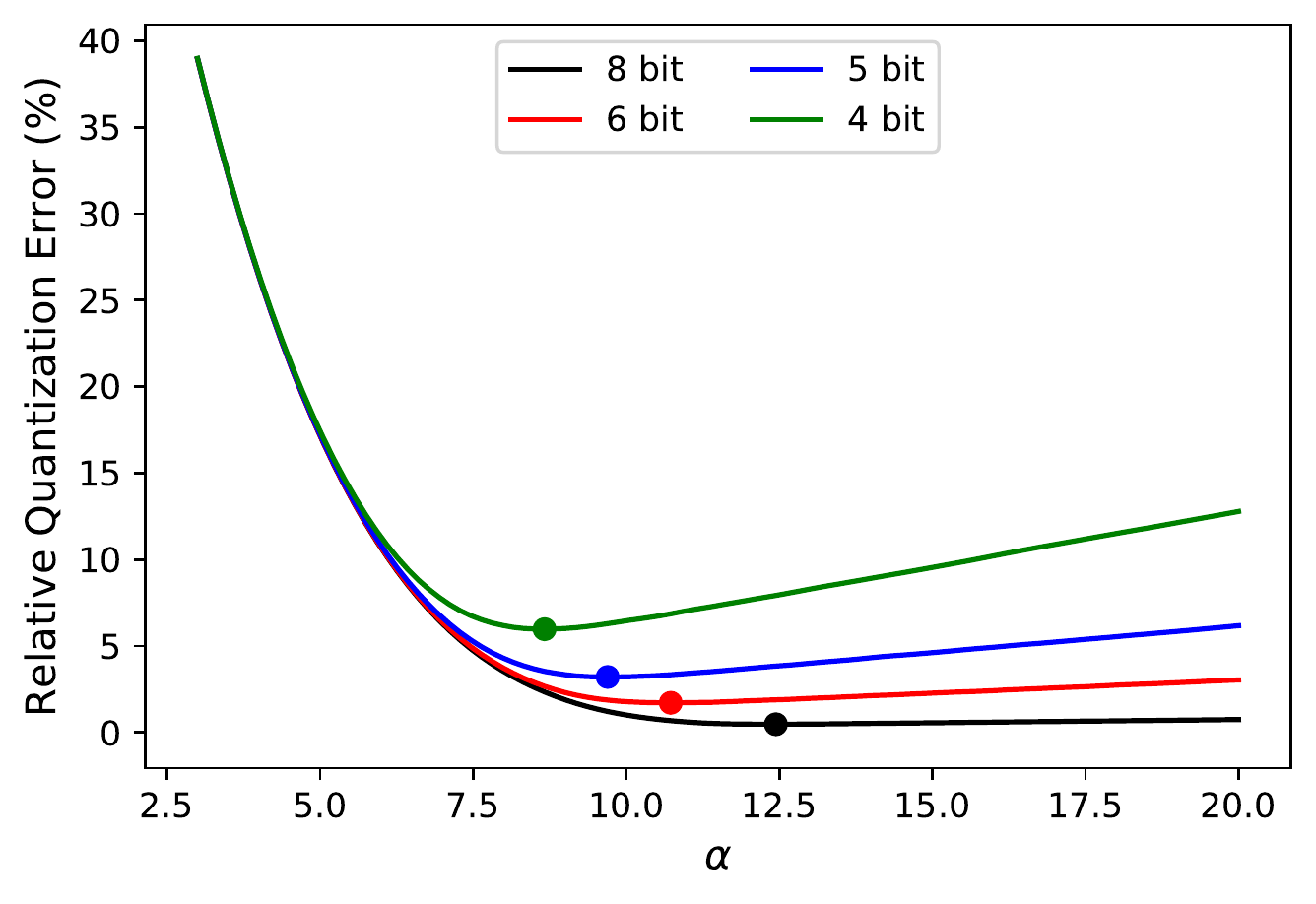}
\end{center}
   \caption{Relationship between relative quantization error and the clipping level $\alpha$ for different bit-widths with a synthetic layer. The dots denote the optimal values of $\alpha$ for least quantization error at different bit-widths.}
\label{fig:rel_quant_err_clipping_levels}
\end{figure}

To understand the underlying mechanism of the degeneration at the lowest bit-width, we plot the clipping levels from different layers in models trained individually with different bit-widths. As shown in Figure~\ref{fig:clipping_levels_sat}, the clipping levels strongly correlates with the bit-width. For the individually trained models, higher bit-widths result in larger values of clipping levels. In the Vanilla Adabits model, the learned clipping levels tend to be smaller than those of high-precision cases, but are larger than those from the model with the lowest bit-width. To understand the relationship between quantization error and clipping levels, we study the characteristics using a synthetic linear layer with 1000 input neurons where the weights are sampled from $\mathcal{N}(0, 1/1000)$ and activations are sampled from a uniform distribution on the interval $[0, 1]$. For each bit-width, the products of the weights and the activations are fed to the $\mathrm{ReLU}$ function to obtain quantized outputs with different clipping levels. The relative error between the full-precision outputs and the quantized outputs are calculated. We plot this quantization error with respect to the clipping levels in Figure~\ref{fig:rel_quant_err_clipping_levels}. It shows that different bit-widths have different behaviors. Quantization error only increases slowly with increase of clipping level for higher bit-width while it increases significantly with increase of clipping level for lower bit-width. Based on the results from Figure~\ref{fig:clipping_levels_sat}, the clipping levels learned by Vanilla Adabits may substantially increase the quantization error for the lowest 4 bit, but do not affect those of the other bit-widths much. Note this is only a qualitative analysis and the quantization errors shown in Figure~\ref{fig:rel_quant_err_clipping_levels} are not proportional to the quantization errors in the trained networks.

\subsubsection{Switchable Clipping Level}

The above observation indicates that facilitating proper clipping levels for each bit-width could be a key factor for optimal performance of AdaBits models. One set of shared clipping levels is difficult, if not impossible, to satisfy requirements from different bit-widths. To this end, we propose a simple treatment to the clipping levels, named \emph{Switchable Clipping Level (S-CL)}, that employs independent clipping levels for different bit-widths in each layer. 
During training of quantized models with adaptive bit-widths, S-CL switch to a corresponding set of clipping levels for each bit-width in all layers. This avoids the clipping level parameters being interfered by other bit-widths, especially undesired quantization error introduced by too large or too small clipping levels for the current bit-width. In this way, the performance degeneration issue on the lowest bit-width using Vanilla Adabits can be alleviated.

The model size is almost unchanged with S-CL, which is a negligible portion of less than $0.1$\textperthousand. For instance, the byte size ratio of clipping levels with other trainable parameters is $0.0246$\textperthousand ~for MobileNet V1, $0.0588$\textperthousand~for MobileNet V2, and $0.0084$\textperthousand~for ResNet50. Meanwhile, S-CL introduces almost no runtime overhead. After re-configuring the model with desired bit-width, it becomes a normal network to run without additional latency and memory cost. These advantages make it a very practical and economical solution to the adaptive bit-widths problem.

\section{Experiments}
\label{sec:experiments}
We evaluate our AdaBits algorithm on the ImageNet classification task and compare the resulted models with those quantized individually with different bit-widths. After that, we analyze the clipping levels in different layers from an AdaBits model. Finally, we give discussion and present some future work.

\subsection{ImageNet Classification}
To examine our proposed methods, we quantize several representative models with adaptive bit-widths, including MobileNet V1/V2 and ResNet50, and evaluate them on the ImageNet dataset, using AdaBits algorithm. 

We follow the same quantization strategy as SAT~\cite{sat}, which first trains a full-precision model, and then uses it as initialization for training the quantized model. The same training hyperparameters and settings are shared between pretraining and finetuning, including initial learning rate, learning rate scheduler, weight decay, the number of epochs, optimizer, batch size, etc. The input images to the model are set to unsigned 8bit integer (uint8), and no standardization (neither demeaning nor normalization) is applied. For the first and last layers, weights are quantized with bit-width of 8~\cite{choi2018pact}, while the input to the last layer is quantized with the same precision as other layers. Meanwhile, bias in the last fully-connected layer(s) and the batch normalization layers are not quantized.

To make a fair comparison, we adopt the same hyper-parameters as SAT~\cite{sat}. The learning rate is initialized to 0.05, and updated every iteration for totally 150 epochs with a cosine learing rate scheduler~\cite{loshchilov2016sgdr} without restart. Parameters are updated by a SGD optimizer, Nesterov momentum with a momentum weight of 0.9 without damping. Weight decay is set to $4\times10^{-5}$. For MobileNet V1/V2, the batch size is set to 2048, while for ResNet50 it is 1024. The warmup strategy suggested in~\cite{goyal2017accurate} is adopted by linearly increasing the learning rate every iteration to a larger value ($\mathrm{batch~size}/256\times0.05$) for the first five epochs before using the cosine annealing scheduler. The input image is randomly cropped to $224\times224$ and randomly flipped horizontally, and is kept as 8 bit unsigned integer with no standardization applied. Besides, we use full-precision models with clamped weight as initial points to finetune quantized models.

\begin{table*}
\begin{center}
\scalebox{.9}{
\begin{tabular}{clccclccc}
\toprule
\multirow{2.5}{*}{Scheme} & \multicolumn{4}{c}{\bf Individual Quantization (SAT)} & \multicolumn{3}{c}{\bf Adaptive Bit-widths} & \multirow{2.5}{*}{BitOPs} \\
\cmidrule(r){2-5}\cmidrule(lr){6-8}
& Name & Bit-width & Size & Top-1 Acc. & Name & Size & Top-1 Acc. & 
\\ \midrule
\multirow{11.8}{*}{Original} & MobileNet V1 & 8 bit & 4.10 MB & 72.6 & \multirow{4}{*}{\shortstack[l]{AB-MobileNet V1 \\ \scalebox{0.8}{$[8, 6, 5, 4]$ bits}}} & \multirow{4}{*}{FP} & 72.4 \textsubscript{(-0.2)} & 36.40 B \\
& MobileNet V1 & 6 bit & 3.34 MB & 72.3 & & & 72.4 \textsubscript{(0.1)} & 20.81 B \\
& MobileNet V1 & 5 bit & 2.96 MB & 71.9 & & & 72.1 \textsubscript{(0.2)} & 14.68 B \\
& MobileNet V1 & 4 bit & 2.58 MB & 71.3 & & & 71.1 \textsubscript{(-0.2)} & 9.67 B \\
\cmidrule[.5pt]{2-9}
& MobileNet V2 & 8 bit & 3.44 MB & 72.5 & \multirow{4}{*}{\shortstack[l]{AB-MobileNet V2 \\ \scalebox{0.8}{$[8, 6, 5, 4]$ bits}}} & \multirow{4}{*}{FP} & 72.6 \textsubscript{(0.1)} & 19.25 B \\
& MobileNet V2 & 6 bit & 2.92 MB & 72.3 & & & 72.4 \textsubscript{(0.1)} & 11.17 B \\
& MobileNet V2 & 5 bit & 2.66 MB & 72.0 & & & 72.1 \textsubscript{(0.1)} & 7.99 B \\
& MobileNet V2 & 4 bit & 2.40 MB & 71.1 & & & 70.8 \textsubscript{(-0.3)} & 5.39 B \\
\cmidrule[.5pt]{2-9}
& ResNet50 & 4 bit & 13.34 MB & 76.3 & \multirow{3}{*}{\shortstack[l]{AB-ResNet50 \\ \scalebox{0.8}{$[4, 3, 2]$ bits}}} & \multirow{3}{*}{FP} & 76.1 \textsubscript{(-0.2)} & 71.81 B \\
& ResNet50 & 3 bit & 10.55 MB & 75.9 & & & 75.8 \textsubscript{(-0.1)} & 43.75 B \\
& ResNet50 & 2 bit & 7.75 MB & 73.3 & & & 73.2 \textsubscript{(-0.1)} & 23.71 B \\
\midrule
\multirow{8.5}{*}{Modified} & MobileNet V1 & 8 bit & 4.10 MB & 72.6 & \multirow{4}{*}{\shortstack[l]{AB-MobileNet V1 \\ \scalebox{0.8}{$[8, 6, 5, 4]$ bits}}} & \multirow{4}{*}{4.35 MB} & 72.3 \textsubscript{(-0.3)} & 36.40 B \\
& MobileNet V1 & 6 bit & 3.34 MB & 72.4 & & & 72.3 \textsubscript{(-0.1)} & 20.81 B \\
& MobileNet V1 & 5 bit & 2.96 MB & 72.2 & & & 72.0 \textsubscript{(-0.2)} & 14.68 B \\
& MobileNet V1 & 4 bit & 2.58 MB & 70.5 & & & 70.4 \textsubscript{(-0.1)} & 9.67 B \\
\cmidrule[.5pt]{2-9}
& MobileNet V2 & 8 bit & 3.44 MB & 72.7 & \multirow{4}{*}{\shortstack[l]{AB-MobileNet V2 \\ \scalebox{0.8}{$[8, 6, 5, 4]$ bits}}} & \multirow{4}{*}{3.83 MB} & 72.3 \textsubscript{(-0.4)} & 19.25 B \\
& MobileNet V2 & 6 bit & 2.92 MB & 72.5 & & & 72.3 \textsubscript{(-0.2)} & 11.17 B \\
& MobileNet V2 & 5 bit & 2.66 MB & 72.1 & & & 72.0 \textsubscript{(-0.1)} & 7.99 B \\
& MobileNet V2 & 4 bit & 2.40 MB & 70.3 & & & 70.3 \textsubscript{(0.0)} & 5.39 B \\
\bottomrule
\end{tabular}
}
\end{center}
\caption{Comparison between individual quantization and AdaBits quantization for top-1 validation accuracy (\%) of MobileNet V1/V2 and ResNet50 on ImageNet. Note that we use two quantization schemes to compare our AdaBits with SAT baseline models where ``original'' denotes the original DoReFa scheme and ``modified'' denote the modified scheme in Eq.~\eqref{eq:quant_modified} which enables producing weights for lower bit-width from the 8-bit model. ``FP'' denotes the full-precision models is needed to recover weights in different bit-widths.}
\label{tab:imagenet}
\vspace{-10pt}
\end{table*}

The results for these models are summarized in Table~\ref{tab:imagenet}, where we list the Top-1 accuracy on ImageNet classification task, together with model size and BitOPs. We show results with the original DoReFa scheme for MobileNet V1/V2 and ResNet50, and results with the modified scheme described in Section~\ref{sec:modified} for MobileNet V1/V2. We do not include results on the modified scheme for ResNet50, since we found the 2-bit model in this setting does not converge. We use a prefix of AB- to indicate model quantized with AdaBits. Result of the SAT approach~\cite{sat} is also reported as reference, which based on our knowledge presents the state-of-the-art performance on model quantization. We find that our method is able to achieve almost the same performance as individual quantization for all models across all bit-widths using original scheme. Compared to progressive quantization with ascending bit-width in Table~\ref{tab:progressive}, AdaBits approach on ResNet50 significantly boost the performance on the lowest 2 bit. Compared to progressive quantization with descending bit-width, AdaBits boost accuracy of $2.2\%$ on 4-bit ResNet50 and $2.2\%$ on 3-bit ResNet50, respectively. Compared to Vanilla AdaBits, our final approach with S-CL increase performance on the lowest 4 bit by $0.3\%$ on MobileNet V1 with the original scheme. For models with the modified scheme, AdaBits also achieve similar performance as individual models. The benefit of the modified scheme is that it allows direct adaptation from higher bit-width to lower bit-width which only requires storing the quantized weights for the highest bit-width to greatly reduce the model size. The AdaBits models with the original scheme still need to store full precision weights in order to produce quantized weights in each bit-width. Our results prove that adaptive bit-width is an additional option for adaptive models, which is able to further improve trade-offs between efficiency and accuracy for deep neural networks.

\subsection{Illustration of clipping levels}
\begin{figure}[h!]
\begin{center}
   \includegraphics[width=0.7\linewidth]{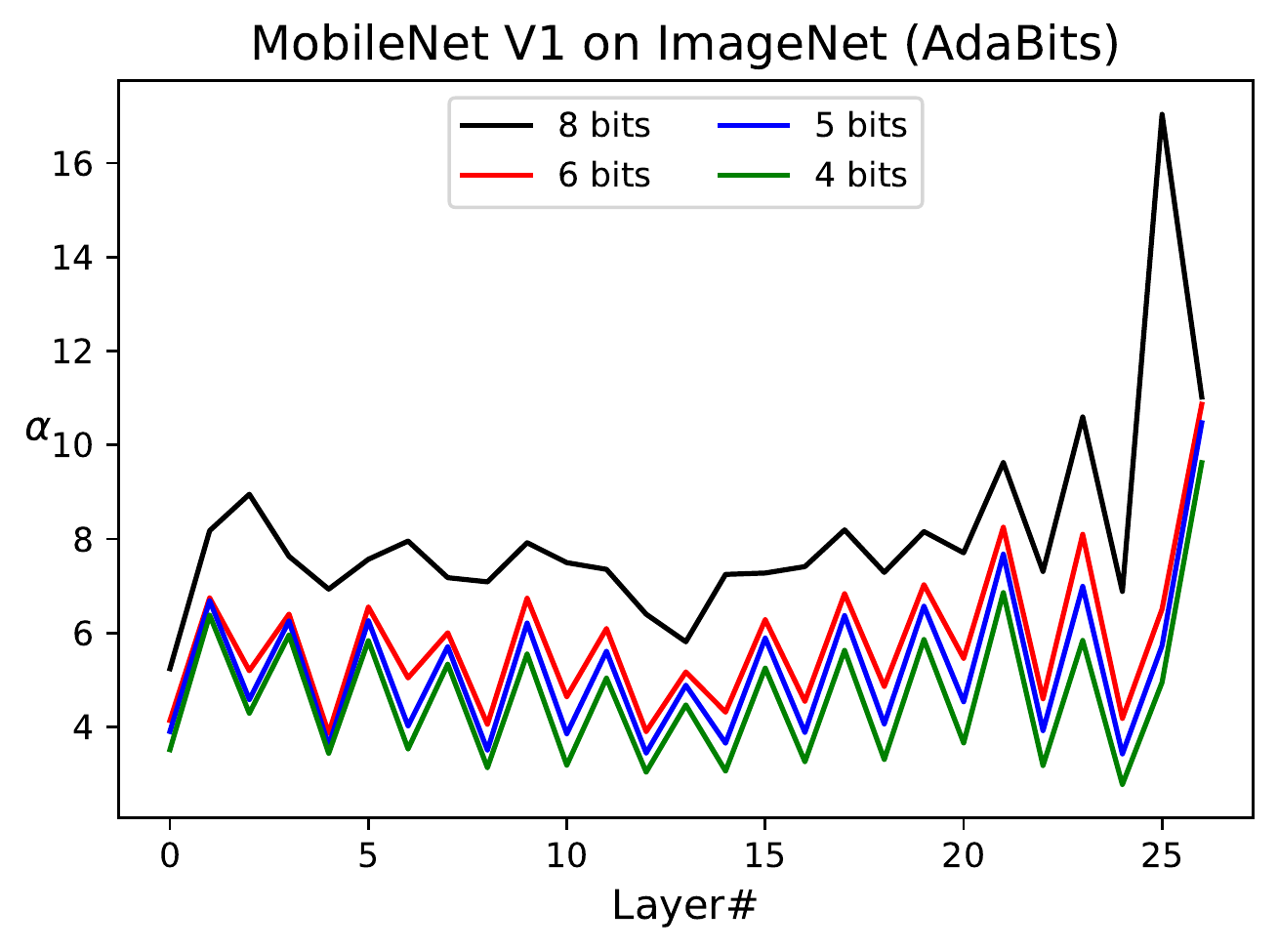}
\end{center}
   \caption{Clipping levels in different layers from AB-MobileNet V1.}
\label{fig:clipping_levels_adabits}
\end{figure}

To understand the impact of S-CL, we visualize the clipping levels from different layers in AB-MobileNet V1 with the original scheme in Figure~\ref{fig:clipping_levels_adabits}. We find different bit-widths indeed lead to different values of clipping levels, which generally follow the order that larger bits have relatively larger clipping levels as in the individual models. By privatizing clipping levels to different bit-widths, different optimal values of clipping levels for different bit-widths can be selected and the optimization of the model can be improved.

\section{Discussion and Future Work}
Our approach for adaptive bit-width indicates that bit-width of quantized models is an additional degree of freedom besides channel number, depth, kernel-size and resolution for adaptive models. Previous work~\cite{sat,autoslim} demonstrate the possibility to utilize the trained adaptive models for neural architecture search algorithms, based on which improved architectures can be discovered under predefined resource constraints. This suggests we might be able to employ the quantized models with adaptive bit-widths to search for bit-width in each layer or channel, for the purpose of mixed-precision quantization~\cite{elthakeb2018releq, wu2018mixed, wang2019haq, uhlich2019differentiable, lou2019autoqb}. On the other hand, adding bit-widths to the list of channel numbers, depth, kernel-size, and resolution enlarges design space for adaptive models, which could enable more powerful adaptive models and facilitate more real world applications, such as face alignment~\cite{miao2018direct,yue2018attentional} and compressive imaging system~\cite{miao2019lambda}.

Evaluation of AdaBits with other quantization methods is another future work. Due to numerous algorithms for neural network quantization, we only select a state-of-the-art algorithm SAT to validate the effectiveness of adaptive bit-width. Since our joint training approach is general and can be combined with any quantization algorithms based on quantization-aware training, we believe similar results can be achieves by  combining other quantization approaches with our  AdaBits algorithm.
\vspace{-5pt}

\section{Conclusion}
\vspace{-5pt}
In this paper, we investigate the possibility to adaptively configure bit-widths for  deep neural networks. After investigating several baseline methods, we propose a joint training approach to optimize all selected bit-widths in the quantized model. Another treatment named Switchable Clipping Level is proposed to privatize clipping level parameters to each bit-width and to eliminate undesired interference between different bit-widths. The final AdaBits approach achieves similar accuracies as models quantized with different bit-widths individually, for a wide range of models including MobileNet V1/V2 and ResNet50 on the ImageNet dataset. This new kind of adaptive  models widen the choices for designing dynamic models which can instantly adapt to different hardwares and resource constraints.

\section{Acknowledgement}
\vspace{-5pt}
The authors would like to appreciate invaluable discussion with Professor Hao Chen from University of California Davis and Professor Yi Ma from University of California Berkeley. They also would like to thank Hongyi Zhang, Yangyue Wan, Xiaochen Lian and Xiaojie Jin from ByteDance Inc., Yingwei Li and Jieru Mei from John Hopkins University, and Chaosheng Dong from University of Pittsburgh for technical discussion.

{\small
\bibliographystyle{ieee_fullname}
\bibliography{egbib}
}

\clearpage

\twocolumn[
  \begin{@twocolumnfalse}
    \maketitle
    \begin{center}
    \textbf{\Large Supplementary Material for ``AdaBits: Neural Network Quantization with Adaptive Bit-Widths''}
    \end{center}
  \end{@twocolumnfalse}
]

\setcounter{section}{0}
\setcounter{equation}{0}

\renewcommand{\theequation}{S\arabic{equation}}
\renewcommand{\thesection}{S\arabic{section}}

\section{Proof of Theorem 1}
\label{sec:supp:proof}

{\noindent\bf Theorem 1} For any $x$ in $[0, 1]$ and any two positive integers $a > b$,
\begin{equation}
\label{eq:quant_modified_proof}
    \lfloor2^ax\rfloor>>(a-b)=\lfloor2^bx\rfloor
\end{equation}

{\noindent\bf Proof} We need to prove that
\begin{equation}
    \bigg\lfloor\frac{\lfloor2^ax\rfloor}{2^{a-b}}\bigg\rfloor=\lfloor2^bx\rfloor
\end{equation}
We first notice that
\begin{equation}
    \frac{\lfloor2^ax\rfloor}{2^{a-b}}\le\frac{2^ax}{2^{a-b}}=2^bx
\end{equation}
so we have
\begin{equation}
\label{eq:ub}
    \bigg\lfloor\frac{\lfloor2^ax\rfloor}{2^{a-b}}\bigg\rfloor\le\lfloor2^bx\rfloor
\end{equation}
due to the non-decreasing monotonicity of the floor function. At the same time, we have
\begin{subequations}
\begin{align}
    \frac{\lfloor2^ax\rfloor}{2^{a-b}}&=\frac{\lfloor2^{a-b}\cdot2^bx\rfloor}{2^{a-b}}\\
    &\ge\frac{\lfloor2^{a-b}\lfloor2^bx\rfloor\rfloor}{2^{a-b}}\\
    &=\frac{2^{a-b}\lfloor2^bx\rfloor}{2^{a-b}}\\
    &=\lfloor2^bx\rfloor
\end{align}
\end{subequations}
where in the penultimate equality we have used the fact that $a > b$. Thus we have
\begin{equation}
\label{eq:lb}
    \bigg\lfloor\frac{\lfloor2^ax\rfloor}{2^{a-b}}\bigg\rfloor\ge\lfloor2^bx\rfloor
\end{equation}

From~\eqref{eq:ub} and~\eqref{eq:lb}, we can get the desired result~\eqref{eq:quant_modified_proof}.

\end{document}